\documentclass[runningheads]{llncs}
\usepackage{graphicx}
\usepackage{subfigure}
\usepackage{bm}
\usepackage{xspace}
\usepackage{multirow}
\usepackage{booktabs}
\usepackage{amsmath}
\usepackage{xcolor}

\newcommand{\ie}{\emph{i.e.,}\xspace}

\newcommand{\ignore}[1]{}

\begin{document}
\title{Improving Multi-Turn Response Selection Models with Complementary Last-Utterance Selection by Instance Weighting}
%
%
\ignore{
\author{Gaole He$^{1,3}$, Junyi Li$^{2}$, Wayne Xin Zhao$^{2,3*}$, Peiju Liu$^{4}$ and Ji-Rong Wen$^{2,3}$}\thanks{$^*$Corresponding author.}
\affiliation{%
 \institution{$^1$School of Information, Renmin University of China}
 \institution{$^2$Gaoling School of Artificial Intelligence, Renmin University of China}
 \institution{$^3$Beijing Key Laboratory of Big Data Management and Analysis Methods}
 \institution{$^4$School of Electronics Engineering and Computer Science, Peking University}
}
\affiliation{%
  \institution{\{hegaole, lijunyi, jrwen\}@ruc.edu.cn, batmanfly@gmail.com, liupage2016@pku.edu.cn}
}}
\author{Kun Zhou\inst{1} \and
Wayne Xin Zhao\inst{2,3}\thanks{Corresponding Author} \and
Yutao Zhu\inst{4} \and 
\\
Ji-Rong Wen\inst{2,3} \and
Jingsong Yu\inst{1}}
\institute{School of Software and Microelectronics, Peking University \and
Gaoling School of Artificial Intelligence, Renmin University of China
\\
\and Beijing Key Laboratory of Big Data Management and Analysis Methods \and
Université de Montréal, Montréal, Québec, Canada
\email{franciszhou@pku.edu.com,batmanfly@gmail.com,yutao.zhu@umontreal.ca,\\jrwen@ruc.edu.cn,yjs@ss.pku.edu.cn}}
\titlerunning{Improving Multi-Turn Response Selection Models by Instance Weighting}
\maketitle              
\begin{abstract}
Open-domain retrieval-based dialogue systems require a considerable amount of training data to learn their parameters. However, in practice, the negative samples of training data are usually selected from an unannotated conversation data set at random. The generated training data is likely to contain noise and affect the performance of the response selection models. To address this difficulty, we consider utilizing the underlying correlation in the data resource itself to derive different kinds of supervision signals and reduce the influence of noisy data. More specially, we consider a main-complementary task pair. The main task (\ie our focus)  selects the correct response given the last utterance and context, and the complementary task selects the last utterance given the response and context. The key point is that the output of the complementary task is used to set instance weights for the main task.
We conduct extensive experiments in two public datasets and obtain significant improvement in both datasets. We also investigate the variant of our approach in multiple aspects, and the results have verified the effectiveness of our approach.

\keywords{Dialog System  \and Instance Weighting \and Noise Reduction.}
\end{abstract}

\section{Introduction}
Recent years have witnessed remarkable progress in retrieval-based open-domain conversation systems~\cite{Li2015ADO,Ji2014AnIR}.
In the past few years, various methods have been proposed for response selection~\cite{Ji2014AnIR,Wu2016SequentialMN,Zhou2018MultiTurnRS,Feng2019LearningAM}. A key problem in response selection is how to measure the matching degree between a conversation context 
and a response candidate. Many efforts have been made to construct an effective matching model with neural architectures~\cite{Wu2016SequentialMN,Zhou2018MultiTurnRS}.

To construct the training data, a widely adopted approach is pairing a positive response with several randomly selected utterances as negative responses, since the labeling of true negative responses is very time-consuming. 
Although such method does not require labeled negative data, it is likely to bring noise during the random sampling process for negative responses. In real-world datasets, a randomly selected response is likely to be ``\emph{false negative}", in which the sampled response can reply to the last-utterance but is considered as a negative response. For example, the general utterance ``OK!" or ``It's great." can safely respond to many conversations.
As shown in existing studies~\cite{Wu2018LearningMM,Lison2017NotAD,Feng2019LearningAM}, the noise from random sampling will severely affect the performance of the matching model.

\begin{figure*}[!t]
  \includegraphics[width=\textwidth]{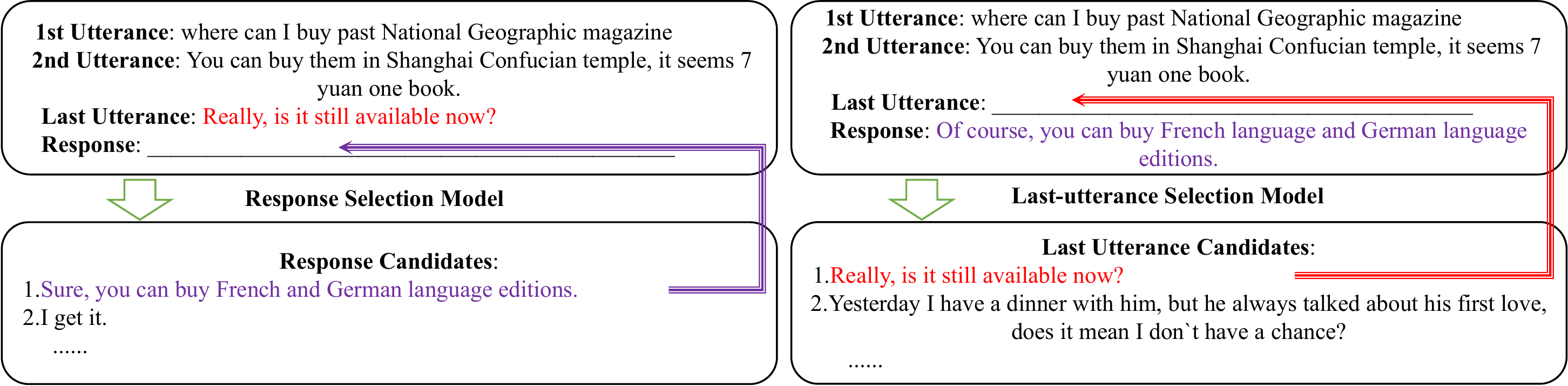}
  \caption{The case of response and last-utterance selection model.} \label{fig-intro}
  \vspace{-0.5cm}
\end{figure*}
However, we do not have any labeled data related to true negative samples. 
To address this difficulty, we find inspiration from the recent progress made in 
complementary learning~\cite{Xia2016DualLF,Wang2018IterativeLW}. 
We design a main-complementary task pair. As shown in Figure~\ref{fig-intro}, the left side is the main task (\ie our focus) which selects the correct response given the last utterance and context, while the right side is the complementary task which selects the last utterance given the response and context.
To implement such a connection, we derive a weighted margin-based optimization objective for the main task. This objective is general to work with various matching models. 
It elegantly utilizes different prospects in utterance selection, either last-utterance selection or response selection. The main task is assisted by the complementary task, and finally, its performance is improved. 

To summarize, the major novelty lies in that the proposed approach can capture different supervision signals from different perspectives, and it is effective to reduce the influence of noisy data.
The approach is general and flexible to apply to various deep matching models. We conduct extensive experiments on two public data sets,
and experimental results on both data sets indicate that the models learned with our approach can significantly outperform their counterparts learned with other strategies.

\section{Related Work}
Recently, data-driven approaches for chatbots~\cite{Serban2015BuildingED,Ji2014AnIR} have achieved promising results. Existing work can be categorized into generation-based methods~\cite{Serban2015BuildingED,vinyals2015neural,Li2015ADO,DBLP:conf/emnlp/ZhouZWLY19} and retrieval-based methods~\cite{Ji2014AnIR,Yan2016LearningTR,zhou2016multi}.The first group of approaches learn response generation from the data. 
Based on the sequence-to-sequence structure with attention mechanism~\cite{vinyals2015neural}, multiple extensions have been made to tackle the ``safe response'' problem and generate informative responses~\cite{Li2015ADO,DBLP:conf/emnlp/ZhouZWLY19}.
The retrieval-based methods try to find the most reasonable response from a large repository of conversational data~\cite{Ji2014AnIR,Wu2016SequentialMN}.
Recent work pays more attention to context-response matching for multi-turn response selection~\cite{Yan2016LearningTR,Wu2016SequentialMN,Zhou2018MultiTurnRS}. 

Instance weighting is a semi-supervised approach proposed by Grandvale et al.~\cite{Grandvalet2004SemisupervisedLB}. The key idea is to utilize weighted margin-based optimization to train the model with a weight function to produce a reward for each instance. Then, researchers used this method to promote the model in noisy training data~\cite{Rebbapragada2007ClassNM}, and extended this method to other tasks~\cite{Jiang2007InstanceWF,Feng2019LearningAM}. 
A recent work showed that the instance weighting strategy can be extended to different machine learning models and validated the improvement in different tasks.

Our work is inspired by the work of using new learning strategies to distinguish the noise in training data~\cite{Shang2018LearningTC,Wu2018LearningMM,Lison2017NotAD}. Shang et al.~\cite{Shang2018LearningTC} and Lison et al.~\cite{Lison2017NotAD} utilized instance weighting strategy in open domain dialog systems via simple methods.
Wu et al.~\cite{Wu2018LearningMM} altered the negative sampling strategy and utilized a sequence-to-sequence model to distinguish false negative samples. 
Feng et al.~\cite{Feng2019LearningAM} proposed three co-teaching mechanisms to reduce noise. 

Different from aforementioned works, we utilize the last-utterance selection task as the complementary task to assist the response selection task by computing the instance weights. This complementary task is similar to the main task since it just exchanges the last utterance with the response. Our method is similar to a dual-learning approach and the difference is that the complementary model is not optimized together with the main model but only provides the instance weights to assist the main task. Besides, the two tasks own the same neural architecture, but leverage different supervision signals from the data. 

\section{Preliminaries}
We denote a conversation as $\{u_1,\cdots,u_j,\cdots,u_n\}$, where each utterance $u_j$ is a conversation sentence.
A dialogue system is built to give the next utterance $u_{n+1}$ to reply $u_n$. 
We refer to the last known utterance (\ie $u_n$) as \emph{last-utterance}, and the utterance to be predicted (\ie $u_{n+1}$) as \emph{response}.

We assume a training set represented by \(\mathcal{D}=\{\langle U_{qi}, q_i, r_i, y_i \rangle \}^{N}_{i=1}\), where $U_{qi}$ denotes the previous utterances $\{u_1,\cdots,u_{n-1}\}$. $q_{i}$ and $r_{i}$ denote the last-utterance and response respectively. $y_i$ is a label indicating whether $r_i$ is an appropriate response to the entire conversation context consisting of $U_{qi}$ and $q_i$.

A retrieval-based dialogue system is designed to select the correct response $r$ from a candidate response pool $\mathcal{R}$ based on the context (namely $U_{q}$ and $q$). This is also commonly called \emph{multi-turn response selection task}~\cite{Yan2016LearningTR,Wu2016SequentialMN}. Formally, we usually solve this task by learning a matching model between last utterance and response given the context to compute the conditional probability of $\text{Pr}(y=1 | q, r, U_q)$, which indicates the probability that $r$ can appropriately reply to $q$.
For simplification, we omit $U_q$ and represent the probability by $\text{Pr}(y=1 | q, r)$.

A commonly adopted loss for the matching model is the Cross-Entropy as:
\begin{equation}
    L_{CE}=-\sum_{i=1}^{N} \big[y_{i}\cdot\log\big(\text{Pr}(y_i | q_i,r_i)\big)+(1-y_{i})\cdot\log\big(1-\text{Pr}(y_i | q_i,r_i)\big)\big].
\label{eq-CE}
\end{equation}
This is indeed a binary classification task. The optimization loss drives the probability of the positive utterance to be one and the negative utterance to be zero.
\section{Approach}
\begin{figure*}[!t]
\small
\includegraphics[width=\textwidth]{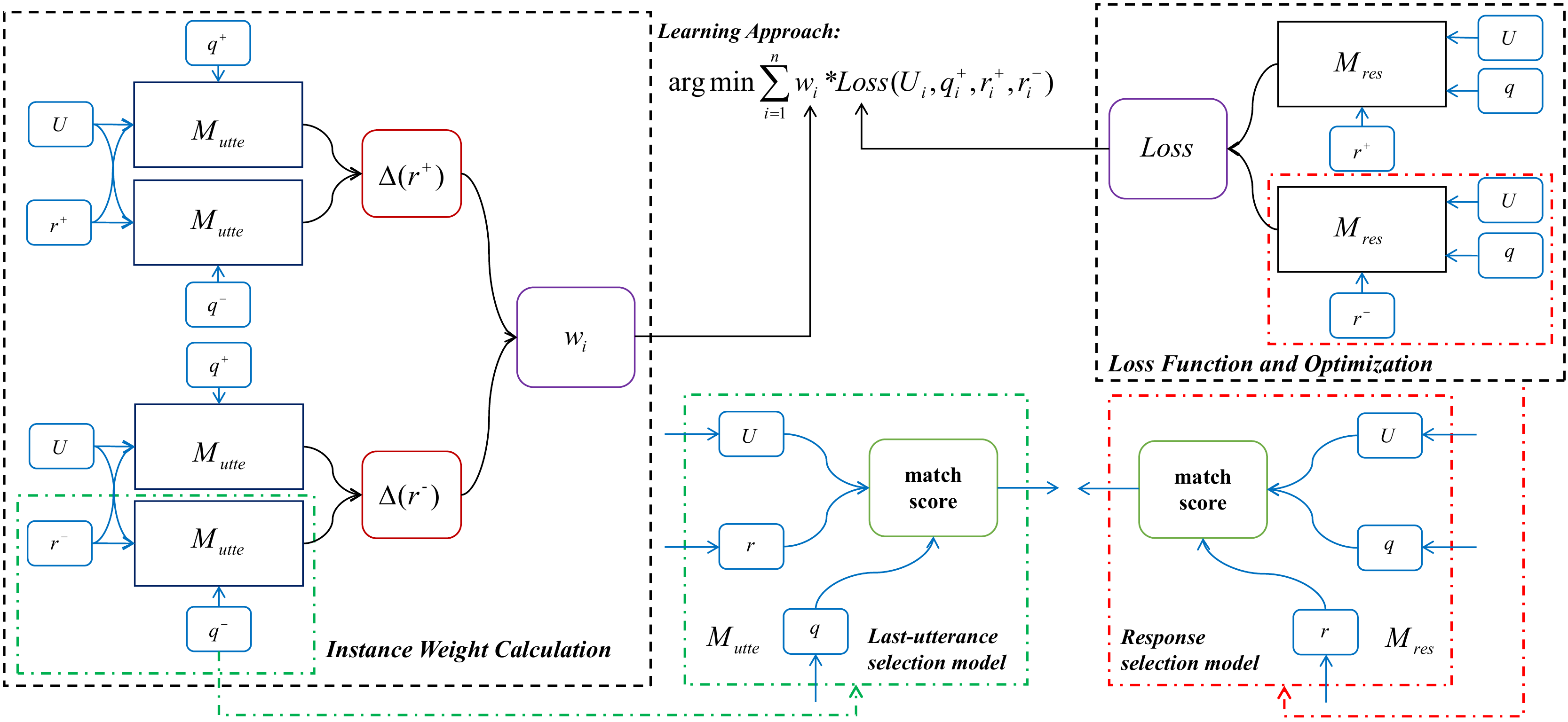}
\caption{The overall sketch of our approach. Our approach contains a main task (Loss Optimization Module) and a complementary task (Instance Weight Calculation Module). Last-utterance selection model $M_{utte}$ is utilized to calculate the instance weight, while response selection model $M_{res}$ is utilized to calculate the loss for optimization.} \label{fig-model}
\vspace{-0.5cm}
\end{figure*}

In this section, we present the  proposed approach to learning matching models for multi-turn response selection. 
Our idea is to assign different weights to training instances, so that we can force the model to focus on confident training instances. An overall illustration of the proposed approach is shown in Figure~\ref{fig-model}.
In our approach, a general weight-enhanced margin-based optimization objective is given, where the weights indicate the reliability level of different instances. We design a complementary task that is to predict last-utterance for automatically setting these weights of training instances used in the main task.   

\subsection{A Pairwise Weight-enhanced Optimization Objective}
Previous methods treat all sampled responses equally, which is easily influenced by the noise in training data.
To address this problem, we propose a general weighted-enhanced optimization objective. 
We consider a pairwise setting: each training instance consists of a positive response and a negative response for a last utterance, denoted by $r^{+}$ and $r^{-}$.
For convenience, we assume each positive response is paired with a single negative sample.

The basic idea is to minimize the \emph{Weighted Margin-based Loss} in a pairwise way, which is defined as: 
\begin{equation}
L_{WM} =  \sum_{i=0}^{N} w_{i} \cdot \max\big\{ \text{Pr}(y=1 | r^{-}_i  , q_i )-\text{Pr}(y=1 | r^{+}_i,  q_i )-\gamma,0\big\},
\label{eq-WM}
\end{equation}
where $w_i$ is the weight for the $i$-instance consisting of $r^{+}_i$ and  $r^{-}_i$. $\gamma \ge 0$ is a parameter to control the threshold of difference. $\text{Pr}(y=1 | r^{+}_i  , q_i )$ and $\text{Pr}(y=1 | r^{-}_i  , q_i )$ denote the conditional probabilities of an utterance being an appropriate and inappropriate response for $q$. When the probability of a negative response is larger than a positive one, we penalize it by summing the difference into the loss. 
This objective is general to work with various matching methods.


\subsection{Instance Weighting with Last-Utterance Selection Model}
A major difficulty in setting weights (shown in Equation~\ref{eq-CE}) is that there is no external supervision information. Inspired by the recent progress made in self-supervised learning and co-teaching~\cite{Feng2019LearningAM,Lison2017NotAD}, we leverage supervision signals from the data itself. Since  response selection aims to select a suitable response from a candidate response pool, we devise a complementary task (\ie last-utterance selection) that is trained with an assistant signal for setting the weights. 

\subsubsection{Last-Utterance Selection}
Similar to response selection, here $q^{-}$ can be sampled negative utterances.  
The complementary task captures data characteristics from a different perspective, so that the learned complementary model can be used to set weights by providing evidence on instance importance. 

\subsubsection{Instance Weighting}
After learning the last-utterance selection model, we now utilize it to set weights for training instances. The basic idea is if an utterance is a proper response, it should well match the real last-utterance $q^{+}$. On the contrary, for a true negative response, it should be uninformative to predict the last-utterance. Therefore, we introduce a new measure $\Delta$ to compute the degree that an utterance is a true positive response as:
\begin{equation}
    \Delta_r=\text{Pr}(y=1| q^{+}, r) -\text{Pr}(y=1 | q^{-}, r),
    \label{eq-delta}
\end{equation}
where $\text{Pr}(y=1| q^{+}, r)$ and $\text{Pr}(y=1 | q^{-}, r)$ are the conditional probabilities of $q^{+}$ and $q^{-}$ learned by the last-utterance selection model. In this way, a false negative response tends to yield a large $\Delta$ value, since it is able to reply to $q^{+}$ and contains useful information to discriminate between  $q^{+}$ and $q^{-}$.
With this measure, we introduce our solution to set the weights defined in Eq.~\ref{eq-WM}.
Recall that a training instance is a pair of positive and ``negative" utterances, and we want to assign a weighted score indicating  how much attention the response selection model should pay.  
Intuitively, a good training instance should be able to provide useful information to discriminate between positive and negative responses. 
 We define the instance weighting formula as:
\begin{equation}
    w_{i}= \min\big\{\max\{\Delta_{r_i^{+}}-\Delta_{r_i^{-}}+\epsilon,0\},1\big\},
    \label{eq-w}
\end{equation}
where $\epsilon$ is a parameter to adjust the mean value of weights, and we constrain the weight $w_i$ to be less equal to 1. From this formula, we can see that a large weight $w_i$ tends to correspond to a large $\Delta_{r_i^{+}}$ (a more informative positive response) and a small $\Delta_{r_i^{-}}$ (a less discriminative negative utterance).

\subsection{Complete Learning Approach and Optimization}
\label{learning-app}
In this part, we present the complete learning approach. 
\subsubsection{Instantiation of the Deep Matching Models}
We instantiate matching models for response selection. Our learning algorithm can work with any deep matching models. Here, we consider two recently proposed attention-based matching models, namely SMN~\cite{Wu2016SequentialMN} and DAM~\cite{Zhou2018MultiTurnRS}.
The SMN model is an RNN-based model. It first constructs semantic representations for context and response by GRU. Then, the matching features are captured by word-level and sequence-level similarity matrix. Finally a convolution neural network is adopted to distill important matching information as a matching vector and an utterance-level GRU is used to compute the matching score.
The DAM model is a deep attention-based model which constructs semantic representation for context and response by a multi-layer transformer. Then, the word-level matching features are captured by cross-attention and self-attention layers. Finally a 3D-convolution is adopted to compute the matching score.
These two models are selected due to their state-of-the-art performance on multi-turn response selection. Besides, previous studies have also adapted them with techniques such as weak-supervised learning~\cite{Wu2016SequentialMN} and co-teach learning~\cite{Feng2019LearningAM}.

\subsubsection{Learning and Optimization}
Given a matching model, we first pre-train it with the cross-entropy in Equation ~\ref{eq-CE}. This step aims to obtain a basic model that will be further fine-tuned by our approach.
For each instance consisting of a positive and a negative response, the last-utterance selection model computes the $\Delta$ value for each response by Equation~\ref{eq-delta}. Then, the weights are derived by Equation~\ref{eq-w} and utilized in the fine-tuning process by Equation~\ref{eq-WM}. The gradient will back-propagate to optimize the parameters in the response selection model (the gradient to last-utterance selection model is obstructed). This training approach encourages the model to focus on more confident instances with the supervision signal from the complementary task. 

\subsubsection{Discussions} In addition to the measure defined in Equation~\ref{eq-w}, we consider using other alternatives to compute $w_{i}$, such as Jaccard similarity and embedding cosine similarity between positive and negative responses. Indeed, it is also possible to replace our multi-turn last-utterance selection model with a single-turn last-utterance selection model to reduce the influence of the context information. Currently, we do not fine-tune the last-utterance selection model, since there is no significant improvement from this strategy in our early experiments. More details will be discussed in Section~\ref{sec-variation}. 

\section{Experiment}
In this section, we first set up the experiments, and then report the results and analysis.
\subsection{Experimental Setup}
\label{sec-experiment setup}

\subsubsection{Construction of the Datasets}
To evaluate the performance of our approach, we use two public open-domain multi-turn conversation datasets. The first dataset is Douban Conversation Corpus (Douban) which is a multi-turn Chinese conversation data set crawled from Douban group\footnote{\url{https://www.douban.com/group/explore}}. This dataset consists of one million context-response pairs for training, 50,000 pairs for validation, and 6,670 pairs for test. Another dataset is E-commerce Dialogue Corpus (ECD)~\cite{Zhang2018ModelingMC}. It consists of real-world conversations between customers and customer service staff in Taobao\footnote{\url{https://www.taobao.com/}}. There are one million context-response pairs in the training set, and 10,000 pairs in both the validation set and the test set. For both datasets, the negative responses in the training set and the validation set are randomly sampled and the ratio of the positive and the negative is 1:1\footnote{In the released training data of ECD, negative ones are automatically collected by ranking the response corpus based on conversation history augmented messages using Apache Lucene. Because retrieval negative samples from the index will bring more noisy data, we reconstruct the negative responses by random sampling from the training data. We also conduct experiments on the original training data and witness less promote than our rebuilt training data.}. In the test set, each context has 10 response candidates retrieved from an index whose appropriateness regarding to the context is judged by human annotators. 

\subsubsection{Task Setting}
We implement our method as~\ref{learning-app}. We select DAM~\cite{Zhou2018MultiTurnRS} and SMN~\cite{Wu2016SequentialMN} as response selection models. We only select DAM~\cite{Zhou2018MultiTurnRS} as our last-utterance selection model not only due to its strong feature extraction ability, but also for guaranteeing the gain only comes from the response selection model. The pre-training process follows the setting in~\cite{Zhou2018MultiTurnRS,Wu2016SequentialMN}. During the instance weighting, we choose 50 as the size of the mini-batches. We use Adam optimizer~\cite{Kingma2014AdamAM} with the learning rate as 1e-4. All gradients are clipped by 1.0 to stabilize the training process. We tune $\gamma$ in \{0,1/8,2/8,3/8,4/8\}, and finally choose 2/8 for Douban dataset, 4/8 for ECD dataset. And we test $\epsilon$ in \{0,1/4,2/4,3/4\}, and find 2/4 is the best choice for both datasets.

Following the works~\cite{Wu2016SequentialMN,Zhou2018MultiTurnRS}, we use \emph{Mean Average Presion} (MAP), \emph{Mean Reciprocal Rank} (MRR) and \emph{Precision at position 1} (P@1) as evaluation metrics.
\subsubsection{Baseline Models}
We combine our approach with SMN and DAM to validate the effect. Besides, we compare our models with a number of baseline models: \\
\textit{SMN}~\cite{Wu2016SequentialMN} and \textit{DAM}~\cite{Zhou2018MultiTurnRS}: We utilize the pre-training results of the two models as baselines to validate the promotion of our proposed method. \\
\textit{Single-turn models}: MV-LSTM~\cite{Wan2016MatchSRNNMT} and match-LSTM~\cite{Wang2015LearningNL} are the typical single-turn matching models. They concatenate all utterances in contexts as a long document for matching. \\
\textit{Multi-view}~\cite{zhou2016multi}: It measures the matching degree between a context and a response candidate in both a word view and an utterance view. \\
\textit{DL2R}~\cite{Yan2016LearningTR}: It represents each utterance in contexts by RNNs and CNNs, and the matching score is computed based on the concatenation of the representations.

In addition to these baseline models, we denote the model with our proposed weighting method as Model-WM. 
\subsection{Results and Analysis}
\begin{table}[!t]
\small
\centering
\caption{Results on two datasets. Numbers marked with * indicate that the improvement is statistically significant compared with the pre-trained baseline (t-test with p-value $<$ 0.05). We copy the numbers from~\cite{Wu2016SequentialMN} for the baseline models. Because the first four baselines obtain similar results in Douban dataset, we only implement two of them in ECD dataset.}
  \begin{tabular}{lcccccc}
    \toprule
     Dataset& \multicolumn{3}{c}{Douban} & \multicolumn{3}{c}{ECD} \\
    \midrule
     Models&MAP &MRR & P@1 &MAP &MRR & P@1\\
    \midrule
    \texttt{MV-LSTM} & 0.498& 0.538& 0.348 &0.613 &0.684 &0.525\\
    \texttt{Match-LSTM} & 0.500& 0.537& 0.345 &- &- &-\\
    \texttt{Multiview} & 0.505& 0.543& 0.342 &- &- &-\\
    \texttt{DL2R} & 0.488& 0.527& 0.330 &0.604 &0.661 &0.489\\
    \midrule
    \texttt{SMN} & 0.530& 0.569& 0.378 &0.666 &0.739 &0.591\\
    \texttt{SMN-WM}& 0.550*& 0.589*& 0.397*&0.670 &0.749* &0.612*\\
    \midrule
    \texttt{DAM} & 0.551& 0.598& 0.423 &0.683 &0.756 &0.621\\
    \texttt{DAM-WM}& \textbf{0.584}*& \textbf{0.636}*& \textbf{0.459}*& \textbf{0.686} &\textbf{0.771}* &\textbf{0.647}*\\
    \bottomrule
  \end{tabular}
  \label{dataset}
  \vspace{-0.5cm}
\end{table}
\label{sec-result}

We present the results of all comparison methods in Table~\ref{dataset}. First, these methods show a consistent trend on both datasets over all metrics, i.e., DAM-WM $>$ DAM $>$ SMN-WM $>$ SMN $>$ other models. We can conclude that DAM and SMN are the best baselines in this task than other models because they can capture more semantic features from word-level and sentence-level matching information. Second, our method yields improvement in SMN and DAM on two datasets, and most of these promotions are statistically significant (t-test with p-value $<$ 0.05). This proves the effectiveness of our instance weighting method. 

Third, the promotion on Douban dataset by our approach is larger than that on ECD dataset. The difference may stem from the distribution of test sets of the two data. The test set of Douban is built from random sampling, while that of the ECD dataset is constructed by a response retrieval system. Therefore, the negative samples are more semantically similar to the positive ones. It is difficult to yield improvement by our approach with SMN and DAM in ECD dataset. Fourth, our method yields less improvement in SMN than DAM. A possible reason is that DAM fits our method better than SMN because DAM is a deep attention-based network, which owns stronger learning capacity. Another possible reason is that DAM is less sensitive to noisy training data since we have observed that the convergence process of SMN is not as stable as DAM.

\subsection{Variations of Our Method}
\begin{table}[!t]
\caption{Evaluation of DAM with different weighting strategies on Douban dataset.}
\small
\centering
  \begin{tabular}{llrrr}
    \toprule
     Method & Models &MAP & MRR & P@1\\
    \midrule
    Original &\texttt{DAM} & 0.551& 0.598 &0.423\\
    \midrule
    \multirow{4}{*}{Heuristic }&\texttt{DAM-uniform} & 0.577& 0.623& 0.433\\
    &\texttt{DAM-random}& 0.549& 0.594& 0.399\\
    &\texttt{DAM-jaccard}& 0.572& 0.622& 0.438\\
    &\texttt{DAM-embedding}& 0.573& 0.615& 0.426\\
    \midrule
    \multirow{3}{*}{Model-based }&\texttt{DAM-DAM}& 0.580& 0.627& 0.438\\
    &\texttt{DAM-last-WM}& 0.578& 0.625& 0.439\\
    &\texttt{DAM-dual} &0.579 &0.621 &0.430\\
    \midrule
    Ours&\texttt{DAM-WM}& 0.584& 0.636& 0.459\\
    \bottomrule
  \end{tabular}
  \vspace{-0.5cm}
  \label{different}
\end{table}
\label{sec-variation}
In this section, we explore a series of variations of our method. 
We replace the multi-turn last-utterance selection with other models or replace the weight produced by Equation~\ref{eq-w} with other heuristic methods. 
In this part, our experiments are conducted on Douban dataset with DAM\cite{Zhou2018MultiTurnRS} as our base model.

\subsubsection{Heuristic Method}
We consider the following methods, which change the weight produced by Equation~\ref{eq-w} with heuristic methods. \\
\textit{DAM-uniform}: we fix the weight as one and follow the same procedure of our learning approach, to validate the effectiveness of our dynamic weight strategy. \\
\textit{DAM-random}: we replace the weight model as a random function to produce random values varied in [0,1]. \\
\textit{DAM-Jaccard}: we use the Jaccard similarity between positive response and negative response as the weight. \\
\textit{DAM-embedding}~\cite{Lison2017NotAD}: we use the cosine similarity between the representation of positive and negative response as the weight. For DAM model, we calculate the average hidden state of all the words in the response as its representation.
\subsubsection{Model-based Method}
We consider the following methods, which change the computing approach of $\Delta$ in Equation~\ref{eq-delta} by substituting our complementary model with other similar models. \\
\textit{DAM-last-WM} replaces the multi-turn last-utterance selection model with a single-turn last-utterance selection model. This method is used to prove the effectiveness of the context information $U$ in the last-utterance selection model.
\textit{DAM-DAM} replaces the last-utterance selection model by a response selection model. We utilize DAM model to produce $Pr(y=1|q^{+},r)$ and $Pr(y=1|q^{-},r)$. \\
\textit{DAM-dual} is a prime-dual approach. The response selection model is the prime model and the last-utterance selection model is the dual model. The two approaches learn instance weights for each other as Equation~\ref{eq-WM}.
\subsubsection{Result Analysis}
Table~\ref{different} reports the results of these different variations of our method on Douban dataset. First, most of these variants outperform DAM model. It demonstrates that these instance weight strategies are effective in noisy data training. Among them, DAM-WM achieves the best results for all the three evaluation metrics. It indicates that our proposed method is more effective. Second, the improvement yielded by heuristic methods is less than model-based methods. A possible reason is that neural networks own stronger semantic capacity and the weights produced by these models can better distinguish noise in training data. Third, heuristic methods achieve worse performance than DAM-uniform. It indicates that Jaccard similarity and cosine similarity of representation are not proper instance weighting functions and bring a negative effect on response selection model. 

Moreover, all these model-based methods receive similar results in all three metrics and outperform DAM model. It indicates that these methods are effective but not as powerful as our proposed method. For DAM-DAM model, a possible reason is that it cannot provide more useful signal for this task than our proposed method. For DAM-last-WM, its last-utterance selection model only utilizes the last utterance therefore it cannot select positive last-utterance confidently\footnote{The last-utterance selection model of DAM-WM obtains 0.846 in P@1 metric while the one of DAM-last-WM only obtains 0.526. The distribution of positive and negative in test data is 1:9}, therefore the distinguish ratio becomes noisy and low confident. For DAM-dual model, we observe that the dual-learning approach does not improve the performance of the last-utterance selection task, the reason may be that the response selection task and last-utterance selection task are not an appropriate dual-task or the dual-learning approach is not proper. We will conduct further investigation to find an appropriate dual-learning approach for this task.

\subsection{Case Study}
\begin{table}[!t]
\centering
\small
\caption{Samples with the maximum and minimum weight learned by our approach. Green checkmarks indicate that the response candidates are proper replies of the contexts from human annotated, while red cross marks indicate inappropriate replies. The first case receives a weight of 0.0 and the negative responses can respond to the contexts to some extent. The second case receives a weight of 1.0 and the negative responses are unrelated to contexts.}
  \begin{tabular}{l|p{5cm}|p{4.6cm}}
    \toprule
    \texttt{Weight} &0.0 &1.0\\
    \midrule
    \texttt{1st Utterance} &Girls shouldn't be too thin, so I gain weight successfully. &You can make a Urban Poster.\\
    \texttt{2nd Utterance} &I am 1.63 meters tall and about 94 kilos, is it too thin? &Nice idea.\\
    \texttt{Last Utterance} &It is just in the right places.&Hello, online celebrity.\\
    \midrule
    \texttt{Pos Response} &I am small boned and look thinner, so the people around me always laugh at me. ( \textcolor{green}{$\surd$} )&I`m not online celebrity. ( \textcolor{green}{$\surd$} )\\
    \texttt{Neg Response} &Haha, I think so.( \textcolor{green}{$\surd$} )&If you carry too many things, please think over again. ( \textcolor{red}{$\times$} )\\
    \bottomrule
  \end{tabular}
  \label{case}
  \vspace{-0.5cm}
\end{table}
Previously, we have shown the effectiveness of our method. In this section, we qualitatively analyze why our method can yield good performance.

We calculate the weights of all the instances in training data of Douban dataset, and select the instances with maximum and minimum weight (1.0 and 0.0) respectively. We present some of them in Table~\ref{case} and annotate them manually. The first case receives a weight of 0.0, which demonstrates that the case is identified as inappropriate negative case by our last-utterance selection model. 
The last case receives a weight of 1.0, and we can identify the positive and negative responses. 
This case study shows that our instance weighting method can identify the false negative samples and punish them with less weight.
\section{Conclusion and Future Work}
Previous studies mainly focus on the neural architecture for multi-turn retrieval-based dialog systems, but neglect the fundamental problem from noisy training data. In this paper, we proposed a novel learning approach that was able to effectively reduce the influence of noisy data. We utilized a complementary task to learn the weights for training instances that were used by the main task. The main task was furthermore fine-tuned according to a weight-enhanced margin-based loss. Such an approach can force the model to focus on more confident training instances. Experimental results on two public datasets have demonstrated the effectiveness of our proposed method.
As future work, we will design other instance weighting methods to detect noise in open domain multi-turn response selection task. Furthermore, we will consider combining our approach with more learning paradigms such as dual-learning and adversarial-learning.

\section*{Acknowledgement}
This work was partially supported by the National Natural Science Foundation of China under Grant No. 61872369 and 61832017, the Fundamental Research Funds for the Central Universities, and Beijing Outstanding Young Scientist Program under Grant No. BJJWZYJH012019100020098, and Beijing Academy of Artificial Intelligence~(BAAI).
%
%
\bibliographystyle{splncs04}
\bibliography{sample-base}
\end{document}